\documentclass[12pt]{article}

\usepackage[nobreak,space,sort,compress]{cite}

\usepackage{sbc-template}
\usepackage{graphicx}
\usepackage[utf8]{inputenc}

\usepackage{amsmath}
\usepackage{enumitem}
\setlist{leftmargin=11mm}
\usepackage{booktabs}
\usepackage[acronym]{glossaries}
\usepackage{xspace}
\usepackage[hyphens]{url}
\usepackage[colorlinks=true,allcolors=black]{hyperref}

\usepackage[table,dvipsnames]{xcolor}

\newcommand\blfootnote[1]{%
  \begingroup
  \renewcommand\thefootnote{}\footnote{#1}%
  \addtocounter{footnote}{-1}%
  \endgroup
}

\newacronymstyle{long-short-br}
{%
  \GlsUseAcrEntryDispStyle{long-short}%
}%
{%
  \GlsUseAcrStyleDefs{long-short}%
}
\setacronymstyle{long-short-br}

\title{Automatic Counting and Identification of Train Wagons\\Based on Computer Vision and Deep Learning}

\author{Rayson Laroca\inst{1}, Alessander Cidral Boslooper\inst{2}, David Menotti\inst{1}}

\address{Department of Informatics, Federal University of Paran\'a, Curitiba, Brazil\\
$^2$Boslooper Railway Technology, Curitiba, Brazil
\email{$^1$\{rblsantos,menotti\}@inf.ufpr.br \quad $^2$alexcidral@hotmail.com}
}
\begin{document} 

\maketitle
\newacronym{cnn}{CNN}{Convolutional Neural Network}
\newacronym{fps}{FPS}{frames per second}
\newacronym{iot}{IoT}{Internet of Things}
\newacronym{mqtt}{MQTT}{Message Queuing Telemetry Transport}
\newacronym{ocr}{OCR}{Optical Character Recognition}
\newacronym{rfid}{RFID}{radio-frequency identification}

\newcommand{\etal}{\textit{et al.}\xspace}
\newcommand{\inpi}{BR512020000808\nobreakdash-9\xspace}
\newcommand{\fps}{16\xspace}
\newcommand{\numimages}{14{,}935\xspace}
\newcommand{\numwagons}{1{,}000\xspace}
\newcommand{\days}{5\xspace}

\newcommand{\counting}{100\xspace}
\newcommand{\identification}{99.7\xspace}
\newcommand{\rejection}{11.6\xspace}
\newcommand{\numvideos}{16\xspace}
\newcommand{\videourl}{\url{https://youtu.be/G2Sb8cGjD5E}}
\newcommand{\urlrgi}{https://cloud.3dissue.net/23598/23629/23815/41039/index.html?page=38}
\glsresetall

\begin{abstract}
In this work, we present a robust and efficient solution for counting and identifying train wagons using computer vision and deep learning.
The proposed solution is cost-effective and can easily replace solutions based on \gls*{rfid}, which are known to have high installation and maintenance costs.
According to our experiments, our two-stage methodology achieves impressive results on real-world scenarios, i.e., 100\% accuracy in the counting stage and 99.7\% recognition rate in the identification one.
Moreover, the system is able to automatically reject some of the train wagons successfully counted, as they have damaged identification codes.
The results achieved were surprising considering that the proposed system requires low processing power (i.e., it can run in low-end setups) and that we used a relatively small number of images to train our \gls*{cnn} for character recognition.
The proposed method is registered, under number \inpi, with the National Institute of Industrial Property~(Brazil).
%
%
\blfootnote{\textit{An article about the proposed system has been published in the October 2020 issue of \href{\urlrgi}{\textbf{Railway Gazette International~\cite{laroca2020visual}}}, the leading business journal for the worldwide rail industry.}}
\end{abstract}

\section{Introduction}

\glsresetall

In rail transport, it is increasingly common the development of systems that enable the automatic counting and identification of wagons (also known as railway cars) during the passage of a train through a station or its entry into a depot~\cite{verma2016automatic,zhang2017convolutional,li2019automatic,liu2019wagon,zou2019image}.
Given its importance, this is a topic that has been addressed in the literature since the mid-1990s~\cite{devena1993number,jang1995twostage}.

Currently, most systems rely on \gls*{rfid} technology for these tasks~\cite{yoon2016automatic,liu2019wagon}.
This method provides fast and accurate results, however, installing extra hardware on each wagon increases both installation and maintenance costs considerably~\cite{yoon2016automatic}.
These high costs were also highlighted in the context of container identification (which is very similar to the identification of wagons) by Verma~\etal~\cite{verma2016automatic}, who stated that although modern containers have spaces reserved for the installation of \gls*{rfid} readers, such readers are not used by any major ocean container carriers due to high costs.

On the other hand, dealing with the automatic counting and identification of train wagons using computer vision has characteristics such as low cost and convenient installation, that is, convenient installation and high maintainability~\cite{li2019automatic,liu2019wagon}.
Such an image-based approach is possible due to the fact that all wagons are identified by an alphanumeric code that contains information about the weight, type and subtype of the wagon~\cite{pavloski2010codigos,abnt2019vagao}.
This code is painted on the front, rear and also on the sides of each wagon, with its exact position in each part varying according to the type and characteristics of the~wagon.

The counting task is challenging due to the fact that each wagon that passes through the camera must be correctly located and tracked so that it is not counted more than once.
The identification task is also difficult since there are no defined parameters for painting the code on the wagon, such as foreground and background colors; type, font size, and character distance, etc.~\cite{zou2019image} (unlike energy meters~\cite{salomon2020deep,laroca2020towards} or license plates~\cite{laroca2018robust,laroca2019efficient}, in which such parameters are well defined). 
Also, the identification code can become dirty and damaged due to prolonged travel in the outdoors~\cite{liu2019wagon}, and the corrugated surface of the wagon can make the projection of the $2$D text slanted and jagged~\cite{verma2016automatic}.

Another point to be taken into account is that the computational approach must be very efficient~\cite{kong2009robust,kumar2007vision}; in other words, the system designed for these tasks should be able to run on low-end setups, including embedded devices, as well as to collect information from every single train passing through the control point where the camera is~installed.

Deep-learning based techniques have been achieving surprising results and overcoming various competitions and machine learning challenges~\cite{krizhevsky2012imagenet,lin2014microsoft,lecun2015deep}, given the ability of such techniques to learn representations or extract features automatically.
Therefore, in this work, we propose a robust and efficient system based on computer vision and deep learning for automatic counting and identification of train wagons that eliminates the high costs related to the use of \gls*{rfid}~technology.

In our experimental evaluation, conducted using Full HD videos, our system achieved impressive results, i.e., $\counting$\% accuracy in the counting stage and a mean recognition rate of $\identification$\% in the identification task.
The proposed approach automatically rejected $\rejection$\% of the train wagons successfully counted, as they have damaged/illegible identification codes.
In a high-end GPU (an NVIDIA Titan Xp), our system is capable of processing \fps \gls*{fps}, which means it could be deployed on embedded devices located on site and still perform its functions in a matter of seconds or~minutes.

The remainder of this paper is organized as follows.
The dataset used in our experiments is described in Section~\ref{sec:dataset}.
The proposed methodology is briefly presented in Section~\ref{sec:proposed}.
We report and discuss the experiments in Section~\ref{sec:experiments}.
Finally, conclusions and future work are presented in Section~\ref{sec:conclusions}.

\section{The Dataset}
\label{sec:dataset}

\begin{figure}[!htb]
    \centering
    \includegraphics[width=0.95\linewidth]{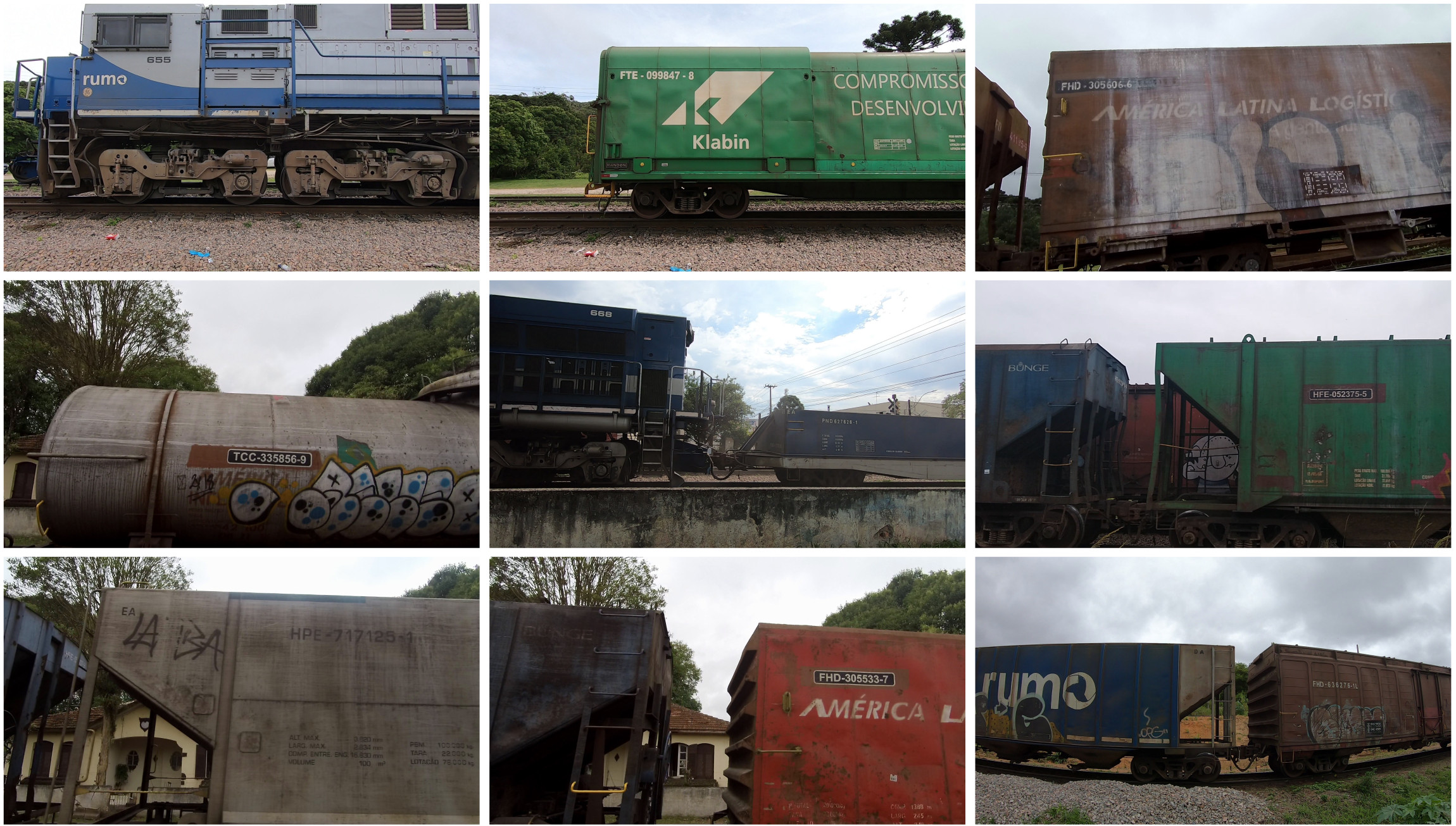}
    
    \vspace{-0.5mm}
    
    \caption{Sample images of the dataset used in this work.}
    \label{fig:samples-dataset}
\end{figure}

For the training and evaluation of our system, we collected $\numimages$ images (extracted from videos recorded at $30$ \gls*{fps}) from approximately $\numwagons$ different wagons.
These images were acquired on five different days in four different locations between November 2019 and February 2020.
Thus, as can be seen in Figure~\ref{fig:samples-dataset}, we captured images of different types of wagons and under different conditions.
It should be noted that, in this work, we use only daytime images for two main reasons: (i)~an infrared camera is required to acquire images at night so that the identification codes are legible; and (ii)~special authorization is required to record videos at certain control points outside business~hours.

Note that (i) the region of interest (i.e., the region containing the identification code) occupies a very small portion of the image; (ii) there are several blocks of text in the wagon (for example, company names and slogans); and that (iii) the distance from the camera to the wagons can vary significantly.
These factors make it difficult to accurately locate the codes in the~images.

The images were acquired with four different cameras in Full HD resolution (i.e., $1{,}920\times1{,}080$ pixels).
As these cameras belong to different price ranges, the images presumably have different levels of quality.
According to Kong \etal~\cite{kong2009robust}, a system for this task must work with non-calibrated cameras (that is, with an arbitrary angle and position) so that it can be deployed anywhere near any train track.
When possible, we collected images of the same train with two different cameras, one on each side of the track, as Liya \& Jilin~\cite{liya2002intelligent}.
In this way, if the code region is damaged on only one side of the wagon, we can still correctly identify the wagon since we have information~redundancy.

\begin{figure}[!htb]
    \centering
    \includegraphics[width=0.7\linewidth]{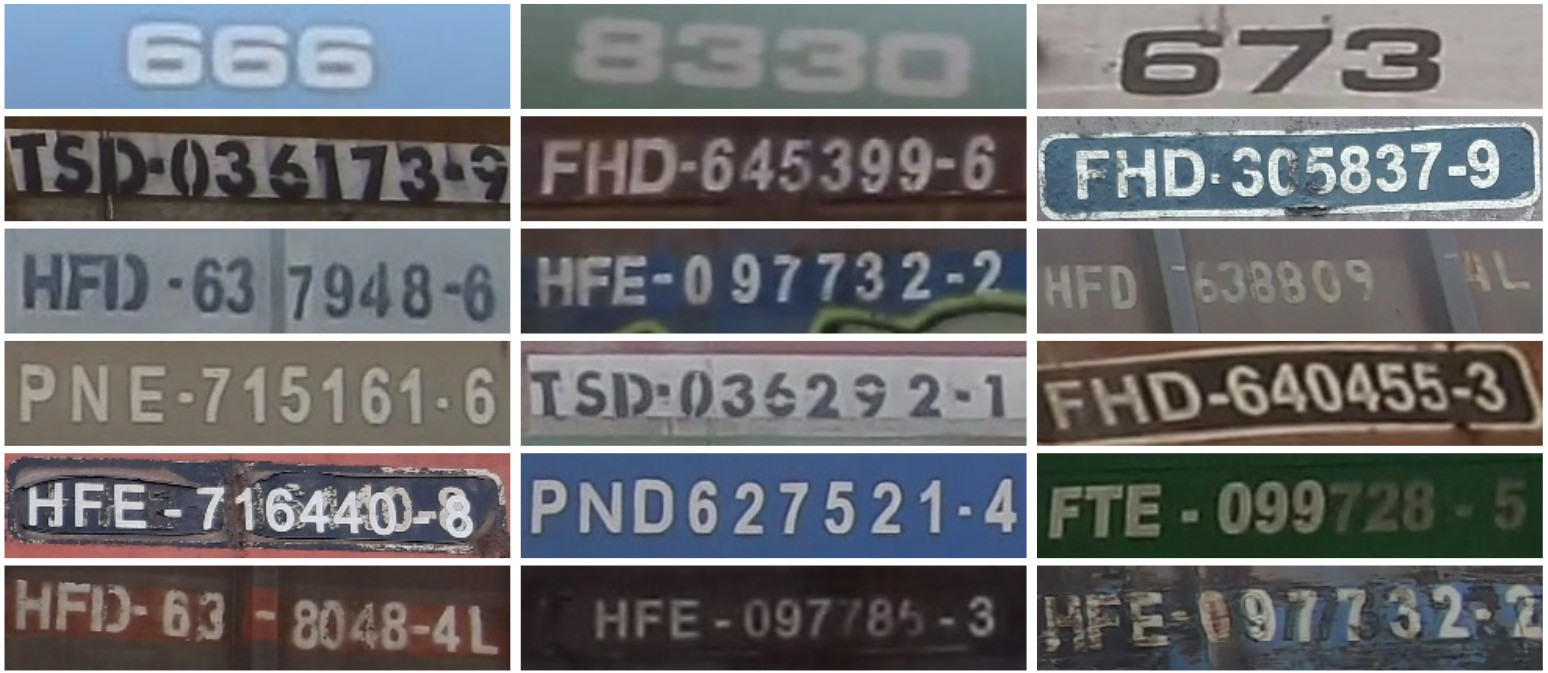}
    
    \vspace{-0.5mm}
    
    \caption{Samples of alphanumeric codes that identify locomotives and wagons.}
    \label{fig:samples-dataset-ocr}
\end{figure}

Figure~\ref{fig:samples-dataset-ocr} shows some characteristic challenges present in the images collected by us, as well as the great variability that exists in the identification codes in different wagons.
As can be seen, the region containing the characters may be or appear corrupted by the presence of noise caused by wear, dirty materials, varied external lighting, etc.
In addition, the spacing between characters can vary considerably, that is, the distance between characters is not uniform, which makes it difficult to segment characters using heuristic-based or attention-based~approaches.

\section{Proposed Approach}
\label{sec:proposed}

This section describes the proposed approach and is divided into two subsections, which are related to the \textit{wagon counting} and \textit{wagon identification} tasks, respectively.

We leverage the high capability of \glspl*{cnn}, the most popular type of deep networks, to handle both tasks.
Note that our system operates fully automatically; in other words, there is no human intervention during the operation of the proposed~method.

In this work, in order not to increase the overall cost of the system, we do not correct possible distortions caused by the camera lenses (according to our experiments, such distortions do not impair the performance of the proposed system), and also do not apply preprocessing techniques to the images before feeding them to the networks.
This same procedure was adopted by Laroca \etal~\cite{laroca2019efficient} and Kumar \etal~\cite{kumar2007vision}.

\subsection{Wagon Counting}

Some approaches in the literature (for example, the one proposed by Liya \& Jilin~\cite{liya2002intelligent}) use different types of sensors to indicate the presence of trains at a certain control point.
Our solution, on the other hand, is based entirely on computer vision.
In this way, implementation and maintenance costs are substantially lower~\cite{yoon2016automatic,verma2016automatic,li2019automatic,liu2019wagon}.

The wagon counting process is based first on the location of regions of interest in the images recorded and then on the analysis and processing of these regions.
More details about this process are not disclosed in this work, as the methodology is registered (under number \inpi) with the National Institute of Industrial Property~(Brazil).

\subsection{Wagon Identification}

In Brazil, the standard coding for the identification of a train wagon is defined by the ABNT NBR 11691:2019 standard~\cite{abnt2019vagao}.
According to this standard, the identification code is composed of exactly $3$ letters and $7$ digits (e.g., HFE-094063-1).
However, there may exist a letter in the eleventh position (e.g., FHD-643258-1L) that indicates the country's region in which the wagon is registered --~we treat this last letter as optional because it was used for many years in previous standards.
The last of the $7$ digits is a check digit, i.e., it is generated by an algorithm based on the previous digits.
Locomotives, on the other hand, are identified by $3$ or $4$ digit codes (for example, 672 and 8330).

In order to identify a wagon, it is sufficient to recognize only the digits since the letters identify characteristics related to the type, subtype, and maximum permissible weight of the wagon~\cite{devena1993number, pavloski2010codigos, abnt2019vagao}.
Nevertheless, in this work, we train a network capable of locating and classifying both letters and digits, so that our system can be easily adapted to identify wagons from other regions/countries where letter recognition is also required.
The network was chosen based on promising results reported in related \gls*{ocr} tasks, such as image-based automatic meter reading~\cite{laroca2019convolutional} and license plate recognition~\cite{oliveira2019vehicle}.
Through heuristic rules, we adapt the results (or predictions) produced by the network according to the patterns described in the previous~paragraph.

Considering that (i)~the region of interest may have been incorrectly located in the previous stage and that (ii)~the wagon identification code may be damaged and, consequently, illegible, we reject the predictions returned by the network in cases where several characters have been predicted with low confidence values and also in cases where the predicted check digit does not correspond with the expected one~\cite{pavloski2010codigos,abnt2019vagao}.

As soon as all wagons of a train pass through a control point (usually a train station or a train depot), the system generates a mosaic that summarizes all information (i.e., the identification code of each wagon and its respective position in the composition) about that~train.
According to Kumar \etal~\cite{kumar2007vision}, the generation of mosaics is important because it results in a large panorama of the train in which all its wagons are visible as a single~block.

The above information is sent to a server/database in the~cloud.
If there are any problems, automatically detected by the system itself, the operator can check the mosaic with the images of the train wagons and, if necessary, make some adjustments to the predictions or even inform other sectors of the railway company about problems in the painting of the identification code (e.g., graffiti) on a given~wagon.
\section{Experiments}
\label{sec:experiments}

In this section, we describe the experiments carried out to verify the robustness and effectiveness of the proposed~method.

\subsection{Setup}
\label{sec:experiments-setup}

All experiments were performed on a computer with an AMD Ryzen Threadripper $1920$X $3.5$GHz CPU, $64$ GB of RAM and an NVIDIA Titan~Xp GPU. 

In the wagon counting stage, all $\numimages$ images collected by us were used either for training or for evaluating the proposed algorithm.
In the experiments related to the wagon identification task, only $1{,}000$ of these images were employed, given the great effort required to manually annotate the bounding box ($x$, $y$, $w$, $h$) of each character in the identification code.
Such an annotation process is necessary for training our \gls*{ocr}~network, which is based on the YOLO object detector~\cite{redmon2016yolo}.

To eliminate any bias in the training process of our system, we adopted a \emph{leave-one-day-out} evaluation protocol, in which we employ images from all days except one for training, while the images from the excluded day are used for~testing.
This procedure is performed for each different day on which we collected images.
In the results section, we report the mean values of accuracy (for the counting task) and recognition rate (for identification) achieved in these $\days$ evaluations (we collected images on $\days$ different~days).

\subsection{Results}

This section reports the results achieved and is divided into two subsections, which are related to the \textit{wagon counting} and \textit{wagon identification} tasks, respectively.

\subsubsection{Wagon Counting}

Through the detection and tracking of the regions of interest (here, the identification codes) in the videos acquired by us, the proposed algorithm achieved $\counting$\% accuracy in this stage, regardless of the camera used and the image/wagon conditions.
To achieve such an impressive result, we explored heuristic rules when processing predictions regarding consecutive frames, ensuring that the number of wagons is correct even if the identification code of a given wagon has not been located.

\begin{figure}[!htb]
    \centering
    \includegraphics[width=0.975\linewidth]{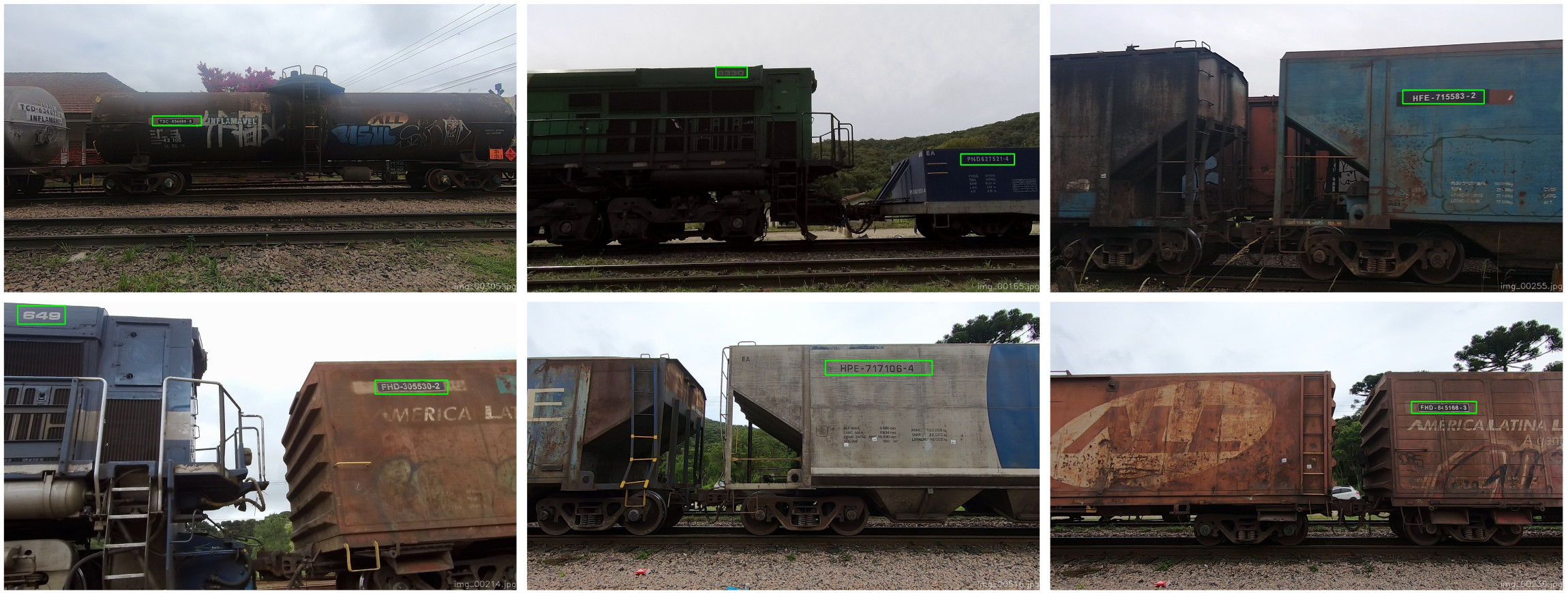}
    
    \vspace{-0.5mm}
    
    \caption{Examples of the detection results obtained for the wagon counting task.}
    \label{fig:roi-detection-results}
\end{figure}

Some detection results are shown in Figure~\ref{fig:roi-detection-results}.
As can be seen, well-located predictions were obtained in wagons of different types, colors and~conditions.
It is worth noting that the number of wagons does not affect the results obtained by the proposed system, as it is capable of efficiently and robustly processing videos of different sizes.
In the videos collected by us, the smallest train composition is composed of $34$ wagons, while the largest is composed of $135$ ($\approx4$ times larger).

\subsubsection{Wagon Identification}

\begin{figure}[!htb]
    \centering
    \includegraphics[width=0.975\linewidth]{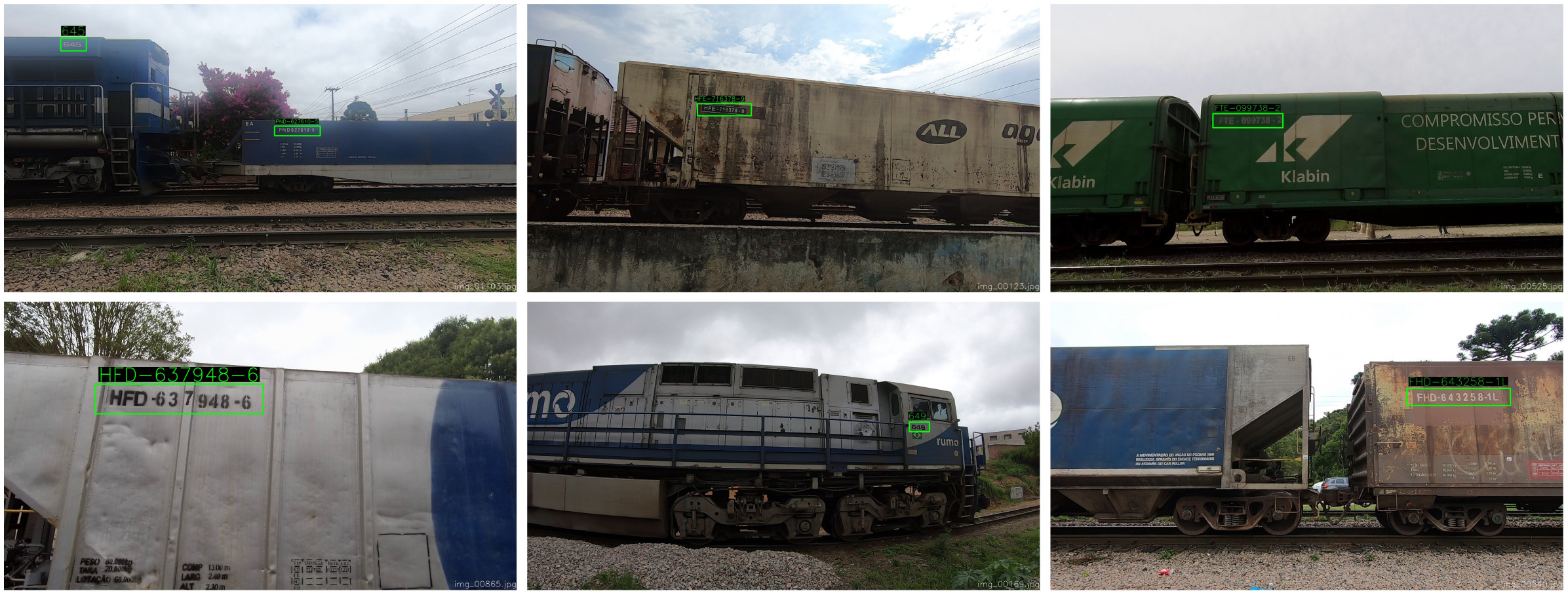}
    
    \vspace{-0.5mm}
    
    \caption{Examples of train wagons correctly identified by the proposed system.}
    \label{fig:ocr-results}
\end{figure}

The results obtained in this stage are shown in Table~\ref{tab:identification-results} (in the second column).
As can be seen, our approach reached an average recognition rate of $97.3$\%, ranging from $94.3$\% to $99.2$\%, based on which videos were used for training and testing.
Such results are surprising considering that we used less than $1{,}000$ images for training the \gls*{ocr} network, as detailed in Section~\ref{sec:experiments-setup}.
According to Figure~\ref{fig:ocr-results}, the proposed method successfully recognized  identification codes, regardless of the color and type of the respective~wagons.

\begin{table}[!htb]
\setlength{\tabcolsep}{12pt}
\centering
\caption{Wagon identification results.}
\label{tab:identification-results}

\vspace{-2.25mm}

\resizebox{0.7\linewidth}{!}{ %
\begin{tabular}{@{}cccc@{}}
\toprule
Day     & \begin{tabular}[c]{@{}c@{}}Recognition Rate\\(Without Rejection)\end{tabular} & Rejection Rate & \begin{tabular}[c]{@{}c@{}}Recognition Rate\\(With Rejection)\end{tabular} \\ \midrule
\# 1    & $95.6$\%   & $15.0$\%    & $\phantom{1}99.0$\%           \\
\# 2    & $98.9$\%   & $11.4$\%    & $100.0$\%          \\
\# 3    & $94.3$\%   & $19.3$\%    & $\phantom{1}99.5$\%           \\
\# 4    & $99.2$\%   & $\phantom{1}4.8$\%     & $100.0$\%          \\
\# 5    & $98.5$\%   & $\phantom{1}7.4$\%     & $100.0$\%          \\ \midrule
Average & $97.3$\%   & $11.6$\%    & $\textbf{\phantom{1}99.7}$\textbf{\%}           \\ \bottomrule
\end{tabular}
} %
\end{table}

\begin{figure}[!htb]
    \centering
    
    \includegraphics[width=0.975\linewidth]{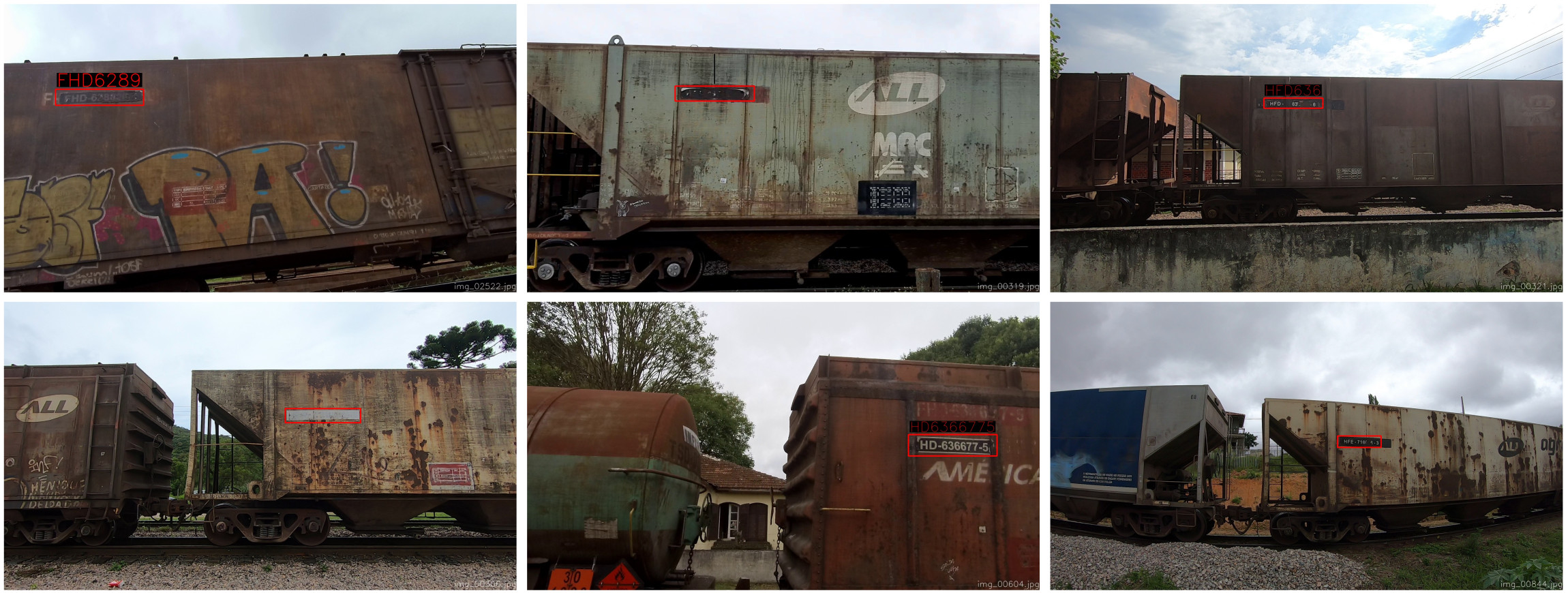}
    
    \vspace{-0.5mm}
    
    \caption{Examples of damaged/illegible identification codes.}
    \label{fig:ocr-results-wrong}
\end{figure}

Regarding the errors, we noticed that they occurred mostly in identification codes with problems in painting.
Some representative examples are shown in Figure~\ref{fig:ocr-results-wrong}.

\begin{figure}[!htb]
    \centering
    \includegraphics[width=0.975\linewidth]{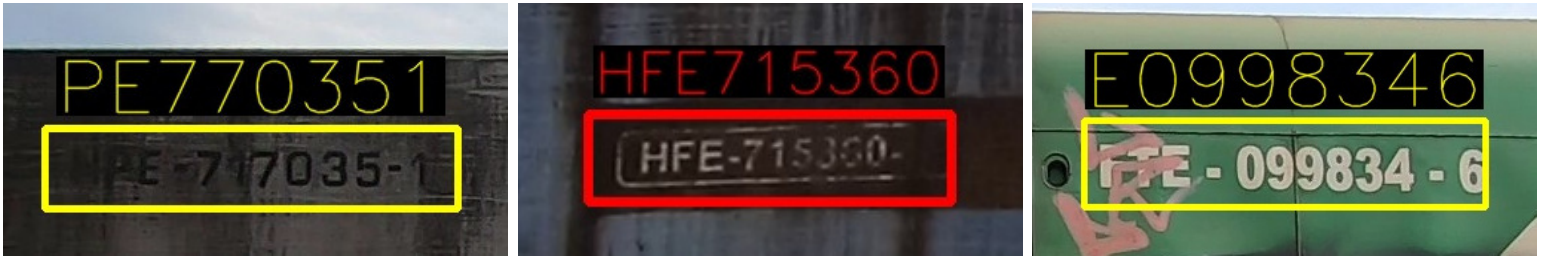}
    
    \vspace{-0.5mm}
    
    \caption{Examples of predictions obtained on damaged identification codes.}
    \label{fig:ocr-results-wrong-cropped}
\end{figure}

As can be seen in Figure~\ref{fig:ocr-results-wrong-cropped}, generally fewer characters are predicted by the \gls*{ocr} network when part of the identification code is damaged or illegible.
For example, in the middle image of Figure~\ref{fig:ocr-results-wrong-cropped} the last character is not visible, while in the rightmost image the first two characters are occluded by graffiti.
It should be noted that such cases are easily rejected taking into account the coding standard defined in~\cite{abnt2019vagao}.

In this sense, we manually labeled all wagons with problems in the identification code as `damaged/illegible' and disregard errors in their codes in a second assessment.
The results can be seen in the last column of Table~\ref{tab:identification-results}.
Considering a rejection rate of $\rejection$\%, our algorithm achieved an impressive recognition rate of $\identification$\%.
Based on these results, the proposed system is very robust and capable of correctly recognizing practically all legible codes, regardless of environmental factors and the distance from the~camera.

It is important to highlight that problems in the painting of the identification codes can be easily solved by the railway companies, both operationally and economically, with a new painting and periodic maintenance in such regions.
Furthermore, through the rejection mechanism, the proposed system automatically identifies which wagons have damaged codes (based on the position of these wagons in the train~composition).

\begin{figure}[!htb]
    \centering
    \includegraphics[width=0.55\linewidth]{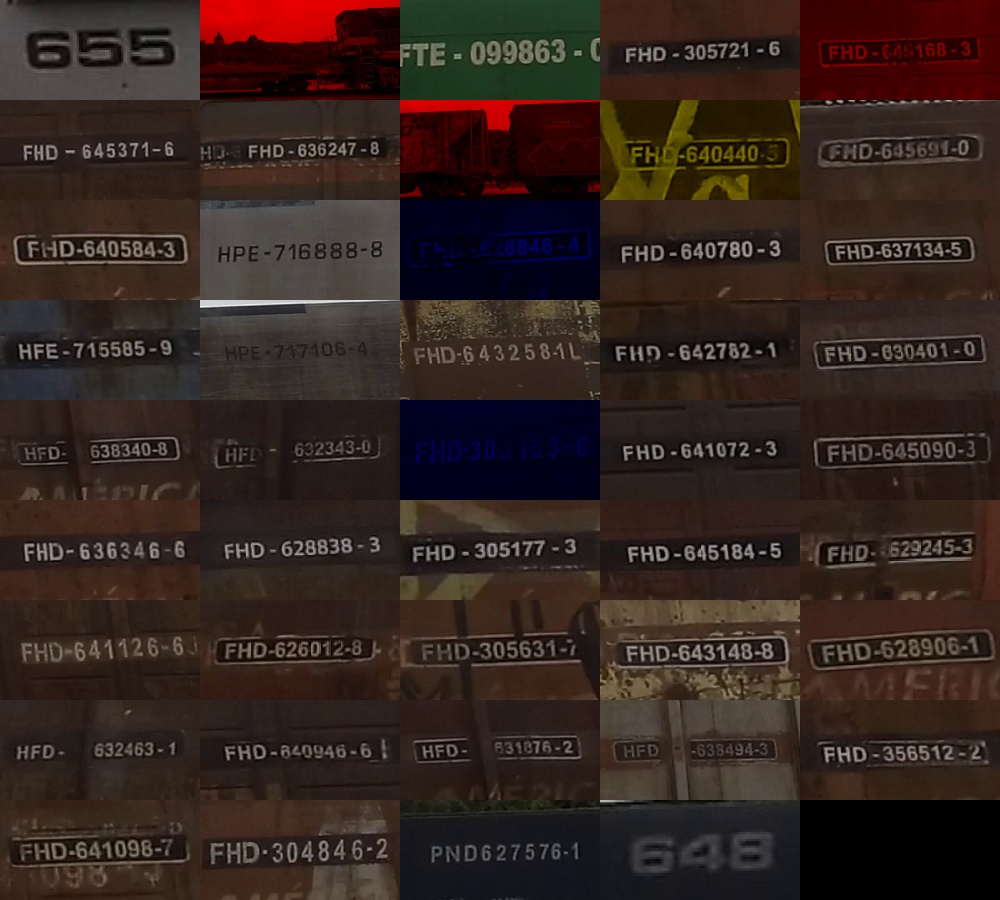}
    
    \vspace{-0.5mm}
    
    \caption{Mosaic generated automatically by the system after a train passes the control point where the camera is installed.}
    \label{fig:mosaic-sample1}
\end{figure}

Figure~\ref{fig:mosaic-sample1} shows a mosaic generated by our system after processing all images of a train that passed through the control point where the camera is installed.
Note that the regions rejected by the system (those shown in red) usually refer to damaged identification codes.
In some situations, it was not possible to detect any region of interest (i.e., identification code); thus, we indicate such cases in the mosaic with a probable frame where the wagon code would be located (e.g., row $2$, column $3$ of Figure~\ref{fig:mosaic-sample1}).
In blue, we show ‘damaged’ codes that were still recognized correctly by the~system.

\begin{figure}[!htb]
    \centering
    \includegraphics[width=0.55\linewidth]{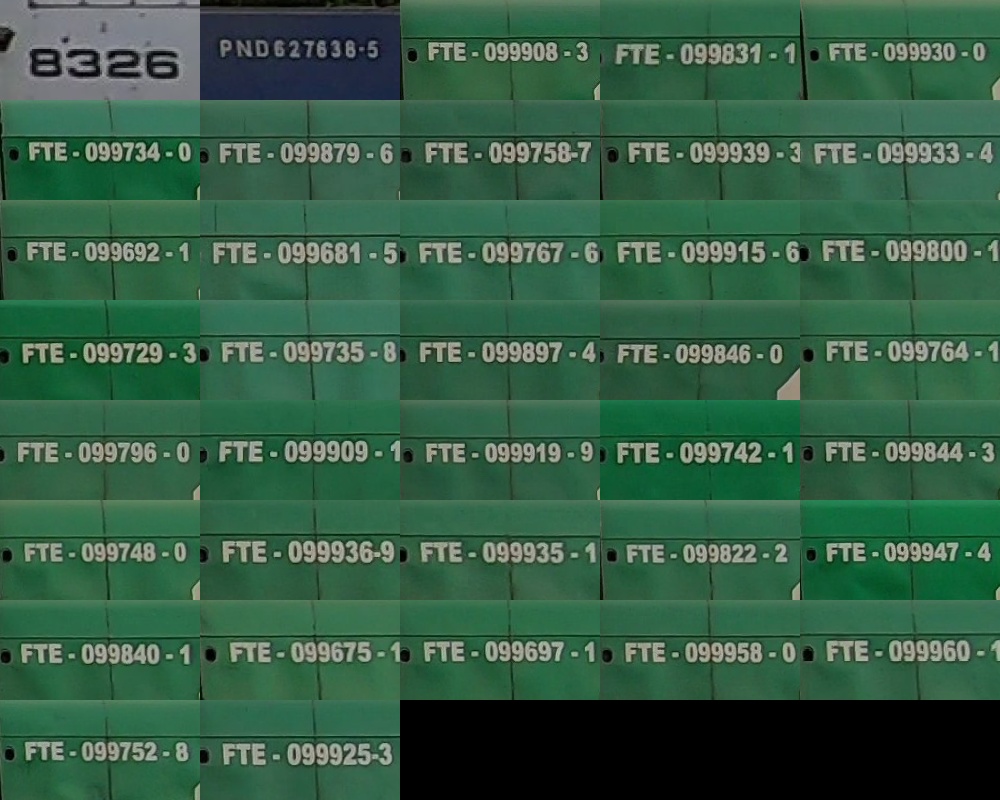}
    
    \vspace{-0.5mm}
    
    \caption{Composition with well-maintained identification codes results in perfect wagon counting and identification.}
    \label{fig:mosaic-sample2}
\end{figure}

We emphasize that the maintenance of the identification codes is essential for the proposed system to obtain optimal results, as illustrated in Figure~\ref{fig:mosaic-sample2}.

\begin{figure}[!htb]
    \centering
    \includegraphics[width=0.48\linewidth]{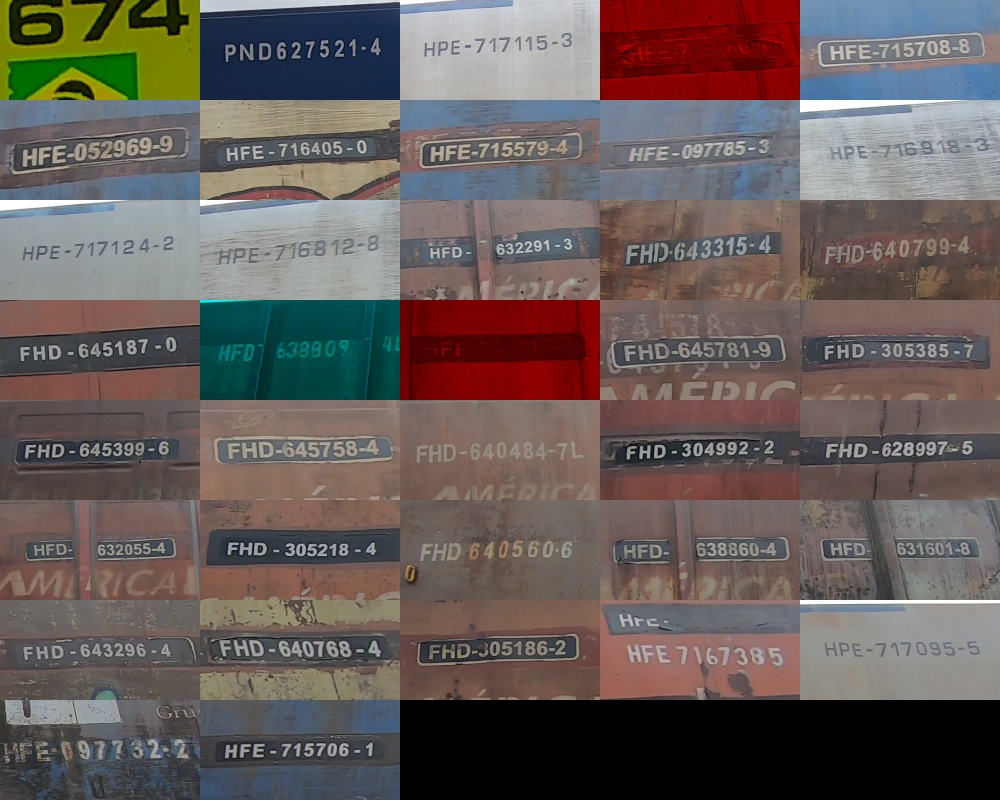}
    \includegraphics[width=0.48\linewidth]{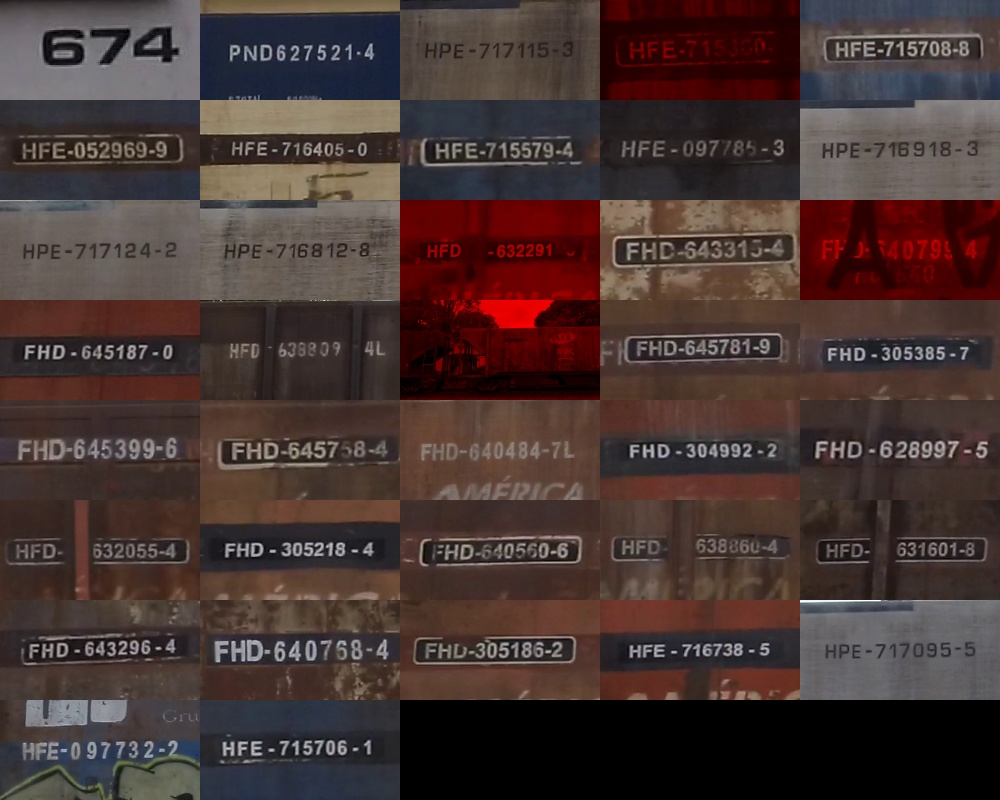}
    
    \vspace{-0.5mm}
    
    \caption{Mosaics of the same train composition recorded by two cameras, one on each side of the track.}
    \label{fig:mosaic-samples34}
\end{figure}

Finally, in Figure~\ref{fig:mosaic-samples34}, we show two mosaics related to a train composition that was recorded by two cameras, one on each side of the track.
Note that the system presented $8$ errors in the recognition of $6$ wagons.
By combining the results (through heuristic rules), the number of errors is reduced to just $2$ wagons since on one side of the wagon the identification code is~recognizable.

\section{Conclusions and Future Work}
\label{sec:conclusions}

In this work, we presented a robust and efficient system based on computer vision and deep learning for automatic counting and identification of train wagons.
Compared with \gls*{rfid}-based methods, the proposed approach is economically advantageous since it has lower installation and maintenance costs.
Our system is registered, under number \inpi, with the National Institute of Industrial Property~(Brazil).

The robustness of the proposed method is remarkable, as it perfectly counted the number of wagons on all videos used in our experiments, and achieved a mean recognition rate of $\identification$\% (considering a rejection rate of $\rejection$\%), even though it was trained in relatively few images and being able to process \fps images per second on a high-end GPU.
We would like to emphasize that our system is likely to become even more robust if more videos (and, consequently, images) are used for training its~networks.

As future work, we intend to carry out experiments in more challenging scenarios, such as images obtained at night, rainy, and foggy periods.
In addition, we plan to ship our solution in the field, carefully defining the best hardware in terms of cost-benefit and also the best position of each camera in order to avoid shadows, reflections and vandalism.
Finally, we want to explore more advanced data augmentation techniques (e.g., those presented in~\cite{ruiz2019anda,ruiz2020ida}) in order to achieve even better results without having to manually label more thousands of images for training our~system.

\section*{Acknowledgments}
This work was supported by the Coordination for the Improvement of Higher Education Personnel~(CAPES) (Social Demand Program) and the National Council for Scientific and Technological Development~(CNPq) (grant numbers~428333/2016-8 and~313423/2017-2).
The Titan~Xp used for this research was donated by the NVIDIA~Corporation.

\clearpage
\bibliographystyle{IEEEtran}
\bibliography{bibtex}

\end{document}